\pdfoutput=1

\documentclass[11pt]{article}
\usepackage{graphicx}
\usepackage{booktabs}
\usepackage{arydshln}
\usepackage{tabularx}
\usepackage{enumitem}
\usepackage[size=tiny]{todonotes}
\usepackage{acl}

\usepackage{times}
\usepackage{latexsym}

\usepackage[T1]{fontenc}
\newcommand{\custompar}[1]{\textbf{#1}}

\newcommand{\sen}[1]{#1}
\newcommand{\se}[1]{#1}
\newcommand{\thd}[1]{\textbf{#1}}

\newcommand{\cotwoeq}{kg CO\textsubscript{2}-eq~}

\usepackage[utf8]{inputenc}
\usepackage{tikz}

\usepackage{microtype}
\usepackage{pgfplots}
\pgfplotsset{width=\textwidth,compat=1.9}
\usepackage{subcaption}
\usepackage{wrapfig}
\usepackage{booktabs}
\usepackage{varioref}
\usepackage{tabularx}

\usepackage{expl3}
\ExplSyntaxOn
\newcommand{\modulo}[2]{\int_mod:nn{#1}{#2}}
\ExplSyntaxOff
\newcommand{\refshape}{circle}
\newcommand{\srcshape}{rectangle}
\newcommand{\chartscat}[5]{
    \ifthenelse{\equal{\modulo{#4}{90}}{0}}{    
        \node[label={[label distance=-2, font=\tiny]#4:#3},#5,fill=#2,inner sep=0pt,minimum size=3pt] at (axis cs: #1) {};
    }{
        \node[label={[label distance=-4, font=\tiny]#4:#3},#5,fill=#2,inner sep=0pt,minimum size=3pt] at (axis cs: #1) {};
    }
}
\newcommand{\chartbase}[5]{\chartscat{#1}{#2}{\textbf{#3}}{#4}{#5}}

\newcommand{\berttiny}{BERT\textsubscript{TINY}}

\newcommand{\bertbase}{BERT\textsubscript{BASE}}

\newcommand{\robertalarge}{RoBERTa\textsubscript{LARGE}}

\newcommand{\tinybertfour}{TinyBERT}

\newcommand{\deebertmnli}{DeeBERT\textsubscript{MNLI}}
\newcommand{\bartbase}{BART\textsubscript{BASE}}
\newcommand{\bartlarge}{BART\textsubscript{LARGE}}
\newcommand{\bartlargecnn}{BART\textsubscript{LARGE}~CNN}
\newcommand{\bartlargepara}{BART\textsubscript{LARGE}~Para}
\newcommand{\dbartsix}{dBART-6-6}
\newcommand{\dbarttwelve}{dBART-12-3}
\newcommand{\dbarttts}{dBART-12-6-t}
\newcommand{\dbartmnli}{dBART-12-9-m}

\newcommand{\xlmrb}{XLM-R\textsubscript{BASE}}
\newcommand{\xlmrl}{XLM-R\textsubscript{LARGE}}
\newcommand{\xdistil}{XtremeDistil}
\newcommand{\mminilmsix}{mMiniLM\textsubscript{6}}
\newcommand{\mminilmtwelve}{mMiniLM\textsubscript{12}}

\title{EffEval: A Comprehensive Evaluation of \\Efficiency for MT Evaluation Metrics}

\author{Daniil Larionov$^1$, Jens Grünwald$^2$, Christoph Leiter$^1$, Steffen Eger$^1$ \\
  $^1$ Department of Computer Science, $^1$ Natural Language Learning Group, Faculty of Technology\\
  $^2$ Technical University of Darmstadt, $^1$ Bielefeld University \\
  \texttt{daniil.larionov@uni-bielefeld.de}
  }

\begin{document}
\maketitle
\begin{abstract}
Efficiency is a key property to foster inclusiveness and reduce environmental costs, especially in an era of LLMs. In this work, we provide a comprehensive evaluation of efficiency for MT evaluation metrics. Our approach involves replacing computation-intensive transformers with lighter alternatives and employing linear and quadratic approximations for alignment algorithms on top of LLM representations. We evaluate six (reference-free and reference-based) metrics across three MT datasets and examine 16 lightweight transformers. In addition, we look into the training efficiency of metrics like COMET by utilizing adapters. Our results indicate that (a) TinyBERT provides the optimal balance between quality and efficiency, (b) CPU speed-ups are more substantial than those on GPU; (c) WMD approximations yield no efficiency gains while reducing quality and (d) adapters enhance training efficiency (regarding backward pass speed and memory requirements) as well as, in some cases, metric quality. These findings can help to strike a balance between evaluation speed and quality, which is essential for effective NLG systems. Furthermore, our research contributes to the ongoing efforts to optimize NLG evaluation metrics with minimal impact on performance. To our knowledge, ours is the most comprehensive analysis of different aspects of efficiency for MT metrics conducted so far.
\end{abstract}

\section{Introduction}

\begin{figure}
    \centering
    \includegraphics[scale=0.5]{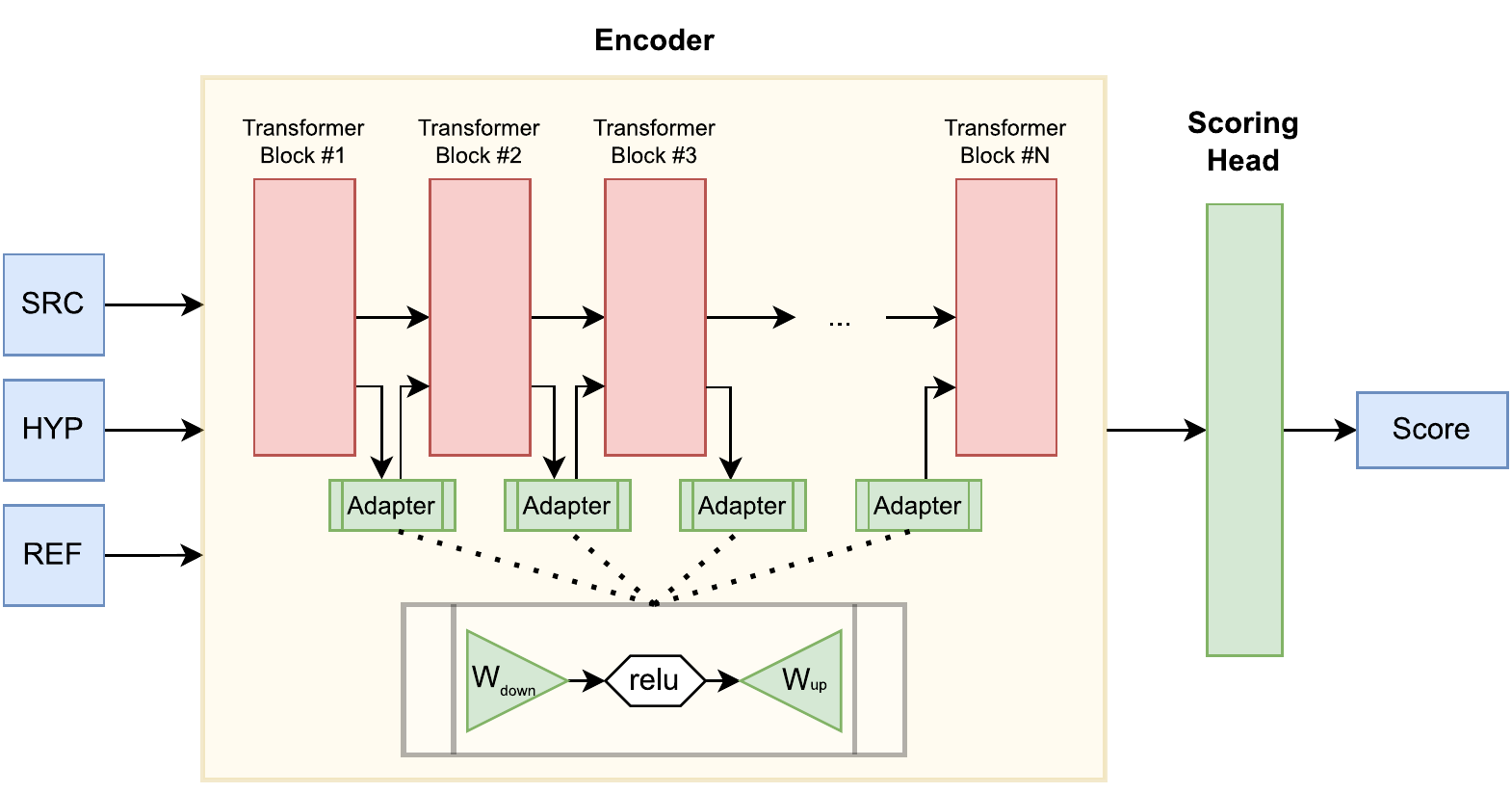}
    \caption{The COMET \se{metric} with \emph{pfeiffer} adapter configuration. Parameters of \textit{red} blocks remain frozen during training, while parameters of \textit{green} blocks are optimized.}
    \label{fig:comet-adapter-pfeiffer}
\end{figure}

Evaluation is crucial to progress in fields such as NLP and machine learning, as it is used to identify and \se{assess} the most promising, state-of-the-art approaches. It is particularly challenging for Natural Language Generation (NLG) systems as text generation is open-ended: multiple outputs, with very different surface-level realizations, can be equally correct \citep{Celikyilmaz2020EvaluationOT}. This insight makes classical lexical overlap metrics such as BLEU~\citep{Papineni2002BleuAM} or ROUGE~\citep{lin-2004-rouge} unsuitable as high-quality evaluation metrics. Consequently, there has been a recent surge of interest~\citep{freitag-etal-2022-results} in developing evaluation metrics based on pretrained large language models (LLMs), which can better cope with lexical variation, thus yielding metrics that correlate much better with human assessments of quality. Notable examples are MoverScore~\citep{zhao_moverscore_2019}, BERTScore~\citep{zhang_bertscore_2020}, and BARTScore~\citep{yuan_bartscore_2021}.

However, basing evaluation metrics on large transformers (and thus boosting the quality of the metrics) also has downsides: for example, the associated computational burden (i) may promote inequality among researchers, hindering diversity, as not everyone has access to expensive compute resources\footnote{E.g., \citet{kamal-eddine-etal-2022-frugalscore} mention that they cannot run variants of BERTScore on a 12GB GPU and the situation would even be worse for more disadvantaged scholars around the world.} and (ii) incurs high environmental costs, one of the most critical issues of our time \citep{strubell_energy_2019}. (iii) Inefficient metrics -- which are in addition non-transparent \citep{Leiter2022TowardsEE} -- may also prevent high-quality metrics from being deployed by the community, a potential reason why older, lower-quality but faster metrics such as BLEU are still popular 
\citep{marie-etal-2021-scientific}.

To better illustrate the issue, let us consider a typical setup for evaluating machine translation~(MT) systems (e.g., the setup of WMT shared tasks). In such a case, one may, for example, have 30k segments to evaluate per language pair, 5 different language pairs, and 50 assessed MT systems. If one uses the BERTScore~\citep{zhang_bertscore_2020} metric with the author-suggested RoBERTa-Large~\citep{liu_roberta_2019} encoder, then it would take 71 hours to completely evaluate all the MT systems on a single Nvidia A100 GPU (given that users do have access to GPUs), producing around 8\cotwoeq of carbon footprint. If access to the GPU is restricted, one could run the metric on a CPU, but it would take more than 950 hours to do a full evaluation, producing 5.4\cotwoeq of carbon footprint. While in the case of a shared task/challenge, the evaluation of model outputs is usually not frequent, there are some cases where evaluation is done constantly, such as hyperparameter search and neural architecture search.

Apart from evaluating MT systems, evaluation metrics have a variety of possible use cases that would greatly benefit from the computationally efficient solutions: {\bf a)} metrics can be used as reward functions in Reinforcement Learning pipelines; {\bf b)} some metrics can be used in the filtering of massive, web-crawled parallel corpora; {\bf c)} they can be used in an online setting for real-time re-ranking of MT systems outputs.

Thus, developing \emph{light-weight high-quality} evaluation metrics for NLG is imperative, which we explore in depth in this work.
We focus on MT as a prime instance of NLG, which also yields a diverse set of scenarios for efficiency, including training efficiency and the efficiency of multilingual models. Nonetheless, we believe that our insights hold more generally.

Our paper about {\bf Eff}icient {\bf Eval}uation~(EffEval) presents the following main contributions:
\begin{itemize}[noitemsep,nolistsep,wide=0pt]
    \item We provide a comprehensive analysis of inducing efficient, high-quality evaluation metrics based on \sen{three} principles: (i) replacing a computation-heavy transformer in the metrics by much smaller ones, obtained e.g.\ via pruning or distillation; (ii) replacing costly alignment techniques (Word Mover Distance; WMD) on top of transformers with cheaper approximations; (iii) implementing parameter-efficient training with adapters.
    \item Our analysis comprises three MT datasets, six evaluation metrics, and 16 light-weight transformers as replacements for the original transformers. 
    \item Based on our large-scale analysis, we find that: (a) for each metric, there is often at least one efficient transformer which leads to higher quality and higher efficiency at the same time, but on average, there is a drop in quality when employing more efficient models; 
    (b) for example, for ``semantic similarity'' metrics like BERTScore, we find that the distilled transformer TinyBERT~\citep{jiao_tinybert_2020} has the best performance-quality tradeoff --- on average, it retains 97\% of the original quality while being 5x faster at inference time;  
    (c) speedups differ substantially on CPU vs.\ GPU; (d) WMD approximations yield no efficiency gains in our experiments (as WMD itself is less costly than embedding computation), but have adverse effects on quality in 2 out of 3 datasets.
    \item Furthermore, we investigate
    training efficiency -- a crucial aspect of recent MT metrics which leverage more and more supervision signals~\citep{rei_comet_2020}, despite criticisms~\citep{belouadi-eger-2023-uscore}
    -- by examining the performance of the popular
    COMET~\citep{rei_comet_2020,rei_comet-22_2022} and COMETINHO~\citep{rei-etal-2022-searching}  trainable metrics when utilizing adapters~\citep{houlsby2019parameter}. Our findings indicate that adapters contribute to an increased backward pass speed by 37\%-102\% and a 26\%-32\%  reduction in memory usage, depending on the model variant. Along with gains on training performance, adapter-enabled models have outperformed the fully-trainable ones, while being trained on the same amount of data.
\end{itemize}
Our code is available at \url{https://github.com/NL2G/effeval}.

\section{Related work}

Our work connects to (1) transformer-based evaluation metrics and to (2) efficiency. Here, we provide only a brief overview of the related papers. Appendix~\ref{appendix:related-work} contains additional related work.

\textbf{Evaluation metrics:} Recent transformer-based metrics utilize BERT-based models like BERTScore~\citep{zhang_bertscore_2020} and MoverScore~\citep{zhao_moverscore_2019}. Extensions include BARTScore~\citep{yuan_bartscore_2021}, which reads off probability estimates as metric scores directly from text generation systems, and MENLI~\citep{Chen2022MENLIRE}, which uses probabilities from models fine-tuned on Natural Language Inference task. These metrics are reference-based (comparing the MT output to a human reference), like BERTScore and MoverScore, or reference-free (comparing the MT output to the source text), like XMoverScore~\citep{zhao-etal-2020-limitations} and SentSim~\citep{song_sentsim_2021}, and some are trained (fine-tuned on human scores) like COMET~\citep{rei_comet_2020} while others are untrained, like BERTScore. Trained metrics typically show higher correlations with human assessments, but require more resources and, thus are more costly. Transversal approaches by~\citet{fu2023gptscore,liu2023gpteval} use LLM predictions.

\textbf{Efficiency:} Techniques like knowledge distillation~\citep{hinton_distilling_2015, ganesh_compressing_2021}, dynamic inference acceleration~\citep{sun_patient_2019, xin_deebert_2020, zhu_leebert_2021}, and adapters~\citep{pfeiffer2020AdapterHub,houlsby2019parameter} seek to improve model efficiency. Knowledge distillation involves a smaller student model learning from a larger teacher, e.g., DistilBERT \citep{sanh_distilbert_2020} and TinyBERT \citep{jiao_tinybert_2020}. \citet{kamal-eddine-etal-2022-frugalscore}
distill an efficient evaluation metric called FrugalScore from the teachers BERTScore/MoverScore. 
Dynamic inference acceleration adds early exit ramps based on representation changes in encoder layers. Adapters freeze pre-trained transformers and train intermediate layers, which can reduce memory usage and improve training speed with varying complexity \citep{pfeiffer_mad-x_2020,he_towards_2022,liu2022fewshot}.
\section{Approach}

For optimizing the metrics, we explore three approaches: (i) We replace transformers with smaller and more efficient variants in \S\ref{sec:build-replace-models} and (ii) we accelerate token matching by calculating Word Centroid Distance~(WCD) and Relaxed Word Mover's Distance~(RWMD) instead of more computationally complex WMD for MoverScore and XMoverScore in \S\ref{sec:build-replace-wmd}. Finally, (iii) we explore the impact of using adapters on training efficiency and metric quality for COMET and COMETINHO.

\subsection{Replacing transformer models}
\label{sec:build-replace-models}
Similarity-based metrics \se{like} BERTScore \citep{zhang_bertscore_2020}, (X)MoverScore~\citep{zhao_moverscore_2019,zhao-etal-2020-limitations}, B\se{ary}Score~\citep{colombo_automatic_2021} and SentSim~\citep{song_sentsim_2021} are not dependent on specific models for calculating token representations. These metrics can leverage any model that generates contextualized vector representations for the input text tokens. By default, \se{the} authors of BERTScore suggest using RoBERTa-Large~\citep{liu_roberta_2019}, which is computationally expensive. To investigate the impact of utilizing more efficient transformer models, we replace default encoders with pruned, distilled, and dynamically accelerated ones:

\textbf{Distillation}: We use DistilBERT \citep{sanh_distilbert_2020} and \tinybertfour{} \citep{jiao_tinybert_2020} for \se{reference-based semantic similarity metrics} BERTScore, MoverScore and BaryScore; dBART \citep{shleifer_pre-trained_2020} for \se{reference-based} BARTScore; 
multilingual DistilMBERT \citep{sanh_distilbert_2020} and \xdistil{} \citep{mukherjee_xtremedistil_2020} for \se{reference-free semantic similarity metrics}  XMoverScore and SentSim; DistilGPT-2 
\citep{von_platen_distilgpt2_2021} for XMoverScore and mMiniLM \citep{wang_minilmv2_2021} for XMoverScore and SentSim.
\textbf{Pruning}: We also examine the performance of one of the miniature BERT models, \berttiny, introduced in~\citep{turc_well-read_2019}, with BERTScore, MoverScore and BaryScore.
\textbf{Dynamic Inference Acceleration}: For BERTScore, MoverScore, and BaryScore, we build a version using DeeBERT's early exiting~\citep{xin_deebert_2020}. 

\subsection{Improving Token Matching Efficiency}\label{sec:build-replace-wmd}
Our second approach for building more efficient metrics involves enhancing the token matching speed in metrics, specifically focusing on WMD in MoverScore \citep{zhao_moverscore_2019,zhao-etal-2020-limitations,colombo_automatic_2021}, a popular approach for token matching in evaluation metrics. Proposed by \citet{kusner_word_2015}, WMD is a specialized version of EMD \citep{rubner_metric_1998} applied to word embeddings. It computes the minimal cost of transforming one document's words into another's while solving a constrained optimization problem with two constraints. However, WMD has a high computational cost due to its cubic inference complexity.

\textbf{WCD}\label{sec:wcd}
\citet{rubner_metric_1998} propose a linear complexity loose lower bound of WMD which \citet{kusner_word_2015} call Word Centroid Distance (WCD). To calculate the distance between documents, WCD first calculates their centroids, i.e.\ the center or average of their word vectors. Then the Euclidean distance between the centroids of these documents is calculated:
\begin{equation}
    \label{eq:wcd}
    \textit{WCD}(x, y) = \sqrt{\frac{1}{|x|}\sum_{i=1}^{|x|}E(x_i) - \frac{1}{|y|}\sum_{j=1}^{|y|}E(y_j)}
\end{equation}
In Eq.~(\ref{eq:wcd}), $x$ and $y$ are 
two documents compared and $E$ is an embedding function.

\textbf{RWMD}\label{sec:rwmd} \citet{kusner_word_2015} also propose the much tighter RWMD, which removes one of the two constraints in the WMD optimization problem. Given the distance of every word in the first document to every word in the other document, RWMD can be calculated with a quadratic complexity.

\subsection{Adapters}\label{sec:adapters}

\textbf{COMET} We incorporate adapters into the training pipeline of the COMET metric to enhance training efficiency by replacing the backbone model from the default pre-trained transformer with its adapter-enabled version.

\textbf{COMETINHO} Furthermore, we apply the same approach to a distillation process described in~\citet{rei-etal-2022-searching}. The training procedure for COMETINHO involves creating a large pseudo-labeled dataset with the help of larger COMET models and training a smaller version based on the smaller MiniLM~\citep{wang2020minilm} pre-trained model (instead of XLM-Roberta-Large~\citep{conneau_unsupervised_2020}) and a smaller estimator layer.

Due to limited hardware resources, we reduce the total training data for COMETINHO. Nevertheless, our goal is not to exactly replicate these models but to investigate whether training metrics can be efficiently obtained without substantial quality loss.
\section{Experimental Setup}

Following related work, we measure the success of our optimized metrics based on the time needed to calculate them, the memory used, and the storage needed to save the program or related data.

\subsection{Evaluation Protocol (disk space, inference time, quality)}

For \textbf{untrained} metrics, we assess efficiency by measuring runtime, memory usage, and model size and compare these with Pearson's $r$ as a quality measure of the metric (correlation with human assessments).

For the \textbf{trainable} COMET metric, we evaluate forward pass and backward pass speeds in tokens per second and memory usage as MB per token. Using relative measures allows us to conduct experiments more efficiently and receive results on the same scale, regardless of batch size and distributed training configuration.
The metric's quality is assessed using Kendall $\tau$ as a correlation with human evaluations. We choose Kendall $\tau$ to make our results more comparable to similar publications for trained metrics such as~\citet{rei-etal-2022-searching}.

\subsection{Untrained metrics}
\paragraph{Runtime}
To calculate the execution/inference time, we measure the timestamp immediately before starting inference and immediately after ending, then report the difference.
    
Since the computing speed of a system depends on factors such as hardware and especially the scheduling of tasks by the operating system, runtimes vary from one run to the next, even when all internal variables stay the same. To reduce this variation, we run every experiment at least three times and average the measured runtimes. For comparability between different metrics, we set the batch size of each metric to 1. We also present an ablation study on the impact of different batch sizes in Appendix~\ref{appendix:batch-size}. It shows that while the ranking of the models in terms of efficiency might differ with higher batch sizes, our main claim (see below: that \tinybertfour~provides the best tradeoff between metric quality and efficiency) still holds.
For comparability between datasets, we divide the total runtime of a metric by the number of segments.

\paragraph{Memory usage, parameters, and disk space}
During inference, we measure the peak memory usage of the program. Since the metrics use the transformers library built on PyTorch, we use PyTorch's \texttt{memory\_stats} array to get the peak usage. We further list the number of parameters in a model and the size needed to save it on a hard drive.

\paragraph{Data}
\label{sec:data}
We use datasets from WMT15, WMT16, and WMT21, all published by the annual Machine Translation conference WMT. The organizers provide source texts, machine translations, and human references. The texts are from various categories, but we only use the ones from \textit{newstest}. The organizers also publish human assessments, which we correlate with our metrics' output as a quality measure. For WMT15 and WMT16, the human scores are Direct Assessment (DA) \citep{stanojevic_results_2015,bojar_results_2016}, while for WMT21, they follow the MQM framework~\citep{lommel2014multidimensional}.

WMT15 provides scored data in 5 language pairs (4 of which are to-English) with 500 segments each, in a total of 2000 segments. WMT16 provides scored data in 7 language pairs (6 of which are to-English) with 560 segments each, in a total of 3360 segments.

For WMT21, scored data is available for three language pairs (en-de, en-ru and zh-en). For the only to-English language pair of those, WMT provides scores for 650 segments, per system. To speed up analysis, we reduce the data to the outputs of only five MT systems (\textit{DIDI-NLP}, \textit{Facebook-AI}, \textit{MiSS}, \textit{NiuTrans} and \textit{SMU}), in total 3250 segments. Appendix~\ref{appendix:wmt-data} provides an overview of language pairs and the number of segments available. 

\paragraph{Hardware}
To get results that are less dependent on our hardware, we consider two different setups:
(i) \textbf{Virtual Machine on a personal computer}: 
The first setup is a virtual machine on a personal computer. The VM has an Intel Core i5-10310U CPU (4 cores, 1.70GHz) and 10 GiB RAM. This setup does not have a GPU that can be used for calculations. 
(ii) \textbf{Compute Cluster}: 
The second setup is the Compute Cluster of a TU Darmstadt Department of Computer Science. 
For our experiments, we chose an Intel Xeon Gold 5218R CPU (20 cores each, 2.10GHz), with a main memory of 64GiB, 8 Nvidia A100 GPUs with 40GB memory each, and runs CentOS Linux 7 as OS. Using the Slurm Workload Manager (version 20.02.2), we limited the usable CPU cores to 4 and GPUs to 1 during our experiments.

For each experiment, we report the runtime on the CPU (as an average of setup 1 and setup 2) and on the GPU (only from setup 2).

\subsection{Trainable metrics}

\textbf{Measuring Training Efficiency} To evaluate the impact of different adapter configurations on training efficiency, we measure both model pass speed and memory usage.

The forward pass speed is assessed by recording timestamps immediately before and after the model forward passes, including the computation of the loss function. The backward pass speed is measured by recording timestamps immediately before and after executing backward pass on the model. The difference between these two timestamps is divided by the total number of tokens in the current minibatch to obtain a normalized speed value. For memory usage measurement, we employ PyTorch memory measurement utilities. Prior to the training step, we reset the current memory usage peak and record the new peak immediately after the step. The final memory usage value is obtained by dividing the memory usage peak by the total number of tokens in the current minibatch.
Although our primary focus in this section is on training efficiency, since inference is covered in another part of the paper, readers can still gain insights into the dynamics of inference efficiency through the measured forward pass speed. This metric provides a useful indicator of the model's performance during the inference stage.

\textbf{Adapter Configurations}
We examine the following adapter configurations and a reference run without adapters.
\begin{itemize}[noitemsep,nolistsep,wide=0pt]
    \item \emph{pfeiffer}~\citep{pfeiffer_mad-x_2020}: A bottleneck adapter layer is placed only after the feedforward block in each transformer layer. The bottleneck consists of down-projection and up-projection with non-linearity in between and an additional residual connection:
    $$
    h \longleftarrow W_{up} \times (W_{down} \times h) + r
    $$
    Here, $W_{down}$ corresponds to a down-projection weight matrix and $W_{up}$ to an up-projection, respectively, and $h$ is an output of the transformer block and $r$ is a residual connection. See Figure~\ref{fig:comet-adapter-pfeiffer} for the architecture of the COMET-like model with \emph{pfeiffer} adapters applied.
    \item \emph{houlsby}~\citep{houlsby2019parameter}: This configuration places the same bottleneck adapter layer both after multi-head attention and the feedforward layer.
    \item \emph{parallel}~\citep{he_towards_2022}: This configuration uses a bottleneck adapter placed in parallel to the transformer layer.
    \item \emph{compacter}~\citep{mahabadi_compacter_2021}: Similar to the bottleneck adapter layer, but instead of linear multiplications, it uses \emph{parametrized hypercomplex multiplication layers} (PHM).
    \item \emph{$(IA)^3$}~\cite{liu2022fewshot}: This configuration introduces trainable vectors $l_W$ into different parts of the transformer model in a way that augments every matrix-multiplication layer with elementwise multiplication:
    $$
    h \longleftarrow l_W \odot (W \times x)
    $$
    Here, $x$ is an output of a transformer block.
\end{itemize}

\paragraph{Data}
For training COMET-based models, we use Direct Assessment (DA) data for the \emph{news} domain from the WMT2020~\citep{barrault-etal-2020-findings} shared task training dataset. In total, it consists of 230,756 segments, covering 14 language pairs. For COMETINHO, we follow the approach of~\citet{rei-etal-2022-searching} and use the data provided by them. We take 10M segments out of the original 45M dataset and compute pseudo-scores using the latest available COMET-22 model\footnote{\url{https://huggingface.co/Unbabel/wmt22-comet-da}}. We train those models in a reference-based setting. Model inputs consist of the source text, its reference translation as well as a hypothesis (which is a machine translation). The model is optimized to predict a score in a range between 0 and 1, where 1 means that the machine translation of the source text is perfect.

We evaluate the metric's quality using WMT21~\citep{akhbardeh-etal-2021-findings} newstest datasets, which consists of 27141 segments: 9750 for \textit{zh-en}, 8959 for \textit{en-de} and 8432 for \textit{en-ru}.

\paragraph{Hardware}
All training-efficiency-related experiments were conducted on Bielefeld's University Compute Cluster. Each node uses an AMD EPYC 7713 64-core Processor, 4 x Nvidia A40, and 512 GB RAM.
\section{Results}

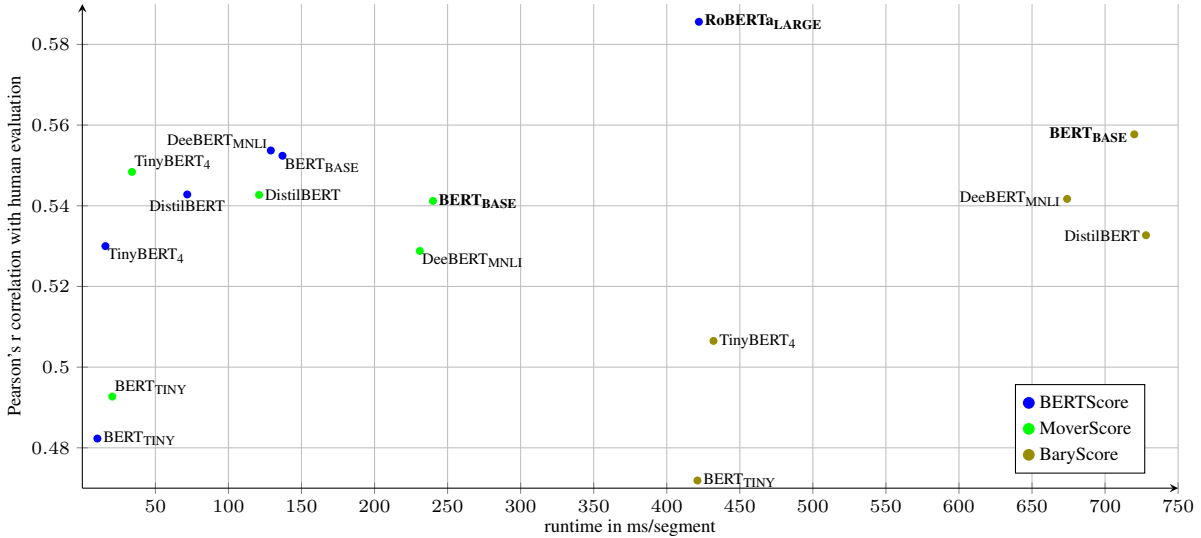
\begin{figure*}[htb]
\centering
    \begin{tikzpicture}
        \scriptsize
        \begin{axis}[
            axis lines=middle,
            xmin=0, xmax=750,
            ymin=0.46, ymax=0.6,
            legend pos = south east,
            legend cell align={left},
            grid = major,
            x label style={at={(axis description cs:0.5,-0.05)},anchor=north},
            y label style={at={(axis description cs:-0.05,0.5)},rotate=90,anchor=south},
            xlabel = runtime in ms/segment,
            ylabel = Pearson's r correlation with human evaluation,
            width=\textwidth, height=0.45\textwidth
        ]
            \addlegendimage{only marks,color=blue}
            \addlegendentry{BERTScore}
            \chartbase{ 422,0.5856}{blue}{\robertalarge}{  0}{\refshape}
            \chartscat{ 137,0.5524}{blue}{\bertbase    }{315}{\refshape}
            \chartscat{10.3,0.4823}{blue}{\berttiny    }{  0}{\refshape}
            \chartscat{71.8,0.5428}{blue}{DistilBERT   }{270}{\refshape}
            \chartscat{15.8,0.5300}{blue}{\tinybertfour}{315}{\refshape}
            \chartscat{ 129,0.5537}{blue}{\deebertmnli }{135}{\refshape}

            \addlegendimage{only marks,color=green}
            \addlegendentry{MoverScore}
            \chartbase{ 240,0.5412}{green}{\bertbase    }{  0}{\refshape}
            \chartscat{20.5,0.4927}{green}{\berttiny    }{ 45}{\refshape}
            \chartscat{ 121,0.5427}{green}{DistilBERT   }{  0}{\refshape}
            \chartscat{34.0,0.5484}{green}{\tinybertfour}{ 45}{\refshape}
            \chartscat{ 231,0.5288}{green}{\deebertmnli }{315}{\refshape}

            \addlegendimage{only marks,color=olive}
            \addlegendentry{BaryScore}
            \chartbase{ 720,0.5577}{olive}{\bertbase    }{180}{\refshape}
            \chartscat{ 421,0.4719}{olive}{\berttiny    }{  0}{\refshape}
            \chartscat{ 728,0.5327}{olive}{DistilBERT   }{180}{\refshape}
            \chartscat{ 432,0.5065}{olive}{\tinybertfour}{  0}{\refshape}
            \chartscat{ 674,0.5417}{olive}{\deebertmnli }{180}{\refshape}
        \end{axis}
    \end{tikzpicture}
    \caption{CPU runtime / correlation plot of BERTScore, MoverScore, BaryScore}
    \label{fig:plot-cpu-runtime-bmb}
\end{figure*}

    We structure this section as follows: We report results for \textbf{untrained} (1) reference-based metrics (\S\ref{sec:ref-based}), (2) reference-free multilingual metrics (\S\ref{sec:xmoverscore-sentsim-word-model}), (3) consider WMD approximations for untrained MoverScore (\S\ref{sec:eva-replace-wmd}) and then (4) we investigate the impact of adapters on training efficiency and metric quality for COMET and COMETINHO \textbf{trained} metrics. 
    Due to space constraints, we relegate details to the appendix~\ref{appendix:experiment-details} and only list the key results as a summary in the main part.

    \subsection{Reference-Based Metrics}\label{sec:ref-based}
    We first consider the ``semantic similarity'' metrics MoverScore, BERTScore, and BaryScore, then consider BARTScore, which is based on text generation and uses different transformer types and corresponding efficient variants.
    
    \subsubsection{
        BERTScore, MoverScore, BaryScore
    }
    \label{sec:eva-replace-models}

    BERTScore, MoverScore, and BaryScore use monolingual BERT-based transformers for embedding references and hypotheses. 
    We replace the embedding model with one pruned model, two distilled models, and an early exiting model. The list of the covered models is as follows: \robertalarge~--- a baseline, the default model for BERTScore; \bertbase~--- the default model for MoverScore and BaryScore;  \berttiny, DistilBERT, \tinybertfour~and 
    \deebertmnli. 
    For detailed experimental setup and results, please refer to Appendix~\S\ref{sec:appendix-eva-replace-models}.\paragraph{Key results} We visualize the runtime of the inference on a CPU and the quality achieved by the metrics in Figure~\ref{fig:plot-cpu-runtime-bmb}. The fastest model for BERTScore, MoverScore, and BaryScore is \berttiny{}, with up to 41x speedup; however, its quality decreases \se{substantially}. \tinybertfour{} achieves a better speedup-quality ratio, with up to 27x speedup while maintaining reasonable quality. Memory measurements are lower for efficient models, with \berttiny{} using 61x and \tinybertfour{} using 18x less memory than the baseline on BERTScore. DistilBERT has a lower speedup, but the quality drop is less compared to \tinybertfour{}. DeeBERT performs similarly to \bertbase{} in quality and efficiency. Further, we observe that speedups --- no matter on which metric, model, or dataset --- are usually not as big on GPU as on CPU, see in Figure~\ref{fig:comparison-cpu-gpu} in Appendix~\ref{appendix:experiment-details}.
    
    
    \subsubsection{BARTScore} 
    \sen{BARTScore} 
    originally uses \bartlarge{} fine-tuned on CNNDM and Parabank2 \citep{yuan_bartscore_2021}. Optimizations of BART were researched by \citet{shleifer_pre-trained_2020}. The models with the best results from them use a \textit{Shrink and Fine-Tune} approach and were also trained on CNNDM. Apart from the original models, we test \bartbase~and several distilled versions --- \dbartsix, \dbarttwelve, \dbarttts~and \dbartmnli. Please refer to Appendix~\S\ref{sec:bartscore-appendix} for detailed setup and results.\paragraph{Key result} \bartbase{} is the fastest and most memory-efficient model but has a substantial
    quality decline. \se{The} distilled version of the BART model, \dbartsix{}, 
    achieves a higher correlation compared to our baseline, \bartlarge, with 1.8x speedup and 1.7x memory efficiency, making it a better choice for quality and efficiency.
    \subsection{Reference-Free Metrics}
    \label{sec:xmoverscore-sentsim-word-model}
    XMoverScore~\citep{zhao-etal-2020-limitations} and SentSim~\citep{song_sentsim_2021} are reference-less metrics,\footnote{When running SentSim, we use the implementation from \citet{belouadi-eger-2023-uscore} since it is better structured and integrates better with our evaluation framework.}
    i.e., they compare hypotheses directly to source texts. Therefore, both of them use a multilingual embedding model since the source and hypothesis are in different languages in MT. 
    \citet{zhao-etal-2020-limitations} realign multilingual embedding spaces on parallel sentences using fast-align \citep{dyer_simple_2013}. We use WikiMatrix~\cite{schwenk-etal-2021-wikimatrix}, instead, to train remappings because it contains more languages. We use the original remappings from \citet{zhao-etal-2020-limitations} for the baseline. 
    We report Pearson's r with (\textit{remap}) and without remapping (\textit{direct}). XMoverScore \se{further} uses a language model for calculating the perplexity of the hypothesis. We also replace this model with a lighter one and measure the efficiency and quality changes. Besides a word level model, SentSim calculates the cosine distance of sentence embeddings \citep{reimers_sentence-bert_2019,reimers_making_2020} of the source and hypothesis sentence. We replace this sentence embedding model with a lighter one. We test 4 models from the `sentence-transformers' model repository: A~(XLM-Roberta) --- a baseline, B~(DistilUSE), C~(MiniLM), and D~(MPNet). For details on the experiment, see Appendix~\S\ref{sec:xmoverscore-sentsim-word-model-appendix}.

    \paragraph{Key results} Both mMiniLMs outperform their baselines in quality and efficiency for XMoverScore and SentSim metrics. \mminilmtwelve{} achieves a 0.032 higher correlation than the baseline on XMoverScore (+8.4\%). Replacing the language model in XMoverScore with DistilGPT-2 results in a 20\% memory reduction and an 8.4\% drop in correlation. For SentSim with sentence embedding models, Model C~(MiniLM) shows the best combination of efficiency and quality. It achieves a 1.5x faster speed on the CPU, uses 1.4x less memory, and only occupies 43\% of the disk space compared to baseline Model A~(XLM-R). Quality drop, in this case, is 7.8\%. 

    \subsection{WMD}
    \label{sec:eva-replace-wmd}
    We replace  WMD in MoverScore and XMoverScore with more efficient variants to speed up 
    runtime. We implement the two variants WCD and RWMD, which have linear and quadratic complexities respectively. See Appendix~\S\ref{sec:wmd-appendix} for more details.

    \paragraph{Key Results} WMD's efficient variants (RWMD and WCD) perform worse in quality for XMoverScore but see an increased quality for MoverScore when using RWMD on WMT21. Their runtime speedup is not substantial due to the time-consuming embedding calculation.

\subsection{Trainable metrics}



\begin{table}[htb]
{\small
    \begin{tabularx}{\linewidth}{l|l|l|l|c}
        \toprule
        \thd{Config} & \thd{Mem.$\downarrow$} & \thd{Fwd.$\uparrow$} & \thd{Bwd.$\uparrow$} & \thd{$\tau\uparrow$}\\
        \midrule
        pfeiffer  & 4.88 & 5123 & {\bf 4808} & 0.273 \\
        parallel  & 4.97 & 5128 & 4525 & {\bf 0.289} \\
        houlsby   & 4.87 & 4607 & 4036 & 0.273 \\
        compacter & {\bf 4.80} & 3649 & 3049 & 0.269 \\
        $(IA)^3$  & 5.76 & {\bf 5195} & 4712 & 0.268 \\
        \midrule
      no adapters & 7.32 & \underline{6247} & 2238 & 0.275 \\
        \midrule
        {\it reference} & - & - & - & {\bf 0.290} \\
        \bottomrule
    \end{tabularx}
    }
    \caption{Training efficiency of COMET models. {\bf Mem.} is the median memory usage in MB per token, {\bf Fwd.} and {\bf Bwd.} are median values of forward pass and backward pass speed respectively, in tokens per second. $\tau$ is the average Kendall $\tau$ across languages in the test set. {\it reference} is the result of applying the latest available COMET-22 model~(\textit{Unbabel/wmt22-comet-da}) through official implementation, both released under Apache 2.0 License}
    \label{tab:comet-scores}
\end{table}

\begin{table}[htb]
{\small
    \begin{tabularx}{\linewidth}{l|l|l|l|c}
        \toprule
        \thd{Config} & \thd{Mem.$\downarrow$} & \thd{Fwd.$\uparrow$} & \thd{Bwd.$\uparrow$} & \thd{$\tau\uparrow$}\\
        \midrule
        pfeiffer  & 0.770 & 25499 & 25774 & 0.252 \\
        parallel  & {\bf 0.741} & 26109 & {\bf 26113} & {\bf 0.252} \\
        houlsby   & 0.769 & 23746 & 21678 & 0.252 \\
        compacter & 0.776 & 18382 & 15671 & 0.243 \\
        $(IA)^3$  & 0.997 & {\bf 27075} & 24804 & 0.248 \\
        \midrule
      no adapters & 1.012 & \underline{31836} & 18941 & 0.243 \\
        \midrule
        {\it reference} & - & - & - & {\bf 0.241} \\
        \bottomrule
    \end{tabularx}
    }
    \caption{Training efficiency of COMETINHO models. {\bf Mem.} is the median memory usage in MB per token, {\bf Fwd.} and {\bf Bwd.} are median values of forward pass and backward pass speed respectively, in tokens per second. $\tau$ is the average Kendall $\tau$ across languages in the test set. {\it reference} is the result of applying the distilled COMET model \emph{eamt22-cometinho-da} \footnote{ \url{https://github.com/Unbabel/COMET/blob/master/MODELS.md}} through \se{the} official implementation.} 
    \label{tab:cometinho-scores}
\end{table}

For both COMET and COMETINHO, 
we evaluate all 5 adapter configurations (\emph{pfeiffer}, \emph{parallel}, \emph{houlsby}, \emph{compacter}, \emph{$(IA)^3$}) + 1 control configuration without adapters. Each configuration is tested three times with different random seeds to minimize the impact of random fluctuations. Our total computational budget for these experiments is approx. 888 GPU-hours: 222 hours of compute time $\times$ 4 A40 GPUs used in parallel.

Results for COMET are presented in Table~\ref{tab:comet-scores}. We observe that adapters can improve metric quality compared to standard training, with the parallel configuration showing a higher average Kendall $\tau$ correlation and almost matching the reference model on 1/4 training data. Although adapters have a slower forward pass speed, increased backward pass speed compensates for it. Lightweight adapter configurations (\emph{pfeiffer} and \emph{parallel}) have higher backward pass speeds than heavyweight compacter, and these adapters also do not show reduced metric quality compared to them.
However, the model with the compacter adapter has the smallest memory footprint of 4.80 MB per token. 


For COMETINHO, similar patterns are observed (Table~\ref{tab:cometinho-scores}). The \emph{parallel} adapters offer the best metric quality and training efficiency, with a 34\% memory reduction and 3.7\%
higher Kendall $\tau$ correlation. This model surpasses the reference model with only 22\% training data.

\section{Analysis \& Discussion}

\begin{table}[!htb]
    \centering
    \begin{tabular}{c|cc}
    \toprule
         \textbf{Model} & \textbf{Quality} & \textbf{Runtime} \\ 
         & & (CPU) \\
         \midrule
         \bertbase & 1.00 & 1.00 \\ 
\berttiny & 0.88 & 8.95 \\
DistilBERT & 0.99& 1.63 \\
TinyBERT & 0.97 & 5.42 \\
\deebertmnli & 0.96 & 1.06 \\ 
    \bottomrule
    \end{tabular}
    \caption{Quality-inference values.}
    \label{table:absolute}
\end{table}

We conduct a deeper analysis on reference-based semantic-similarity metrics BERTScore, MoverScore and BaryScore, all of which use the same efficient encoder architectures (we remove RoBERTa-Large from the analysis, as it has only been used with BERTScore). We compute results individually across the three metrics and the three WMT datasets (then report individual results or averages).

\paragraph{Which speedups are obtained?}
Table~\ref{table:absolute} shows the absolute speedups of encoders and their relative performance deterioration relative to BERT. On a CPU, TinyBERT is 5.4x faster at interference while retaining 97\% of the quality of the original metric, which yields the best tradeoff of quality and inference time. \berttiny{} yields almost 9x faster inference but at the cost of 12 points of performance deterioration.

To illustrate in absolute values, BERTScore with RoBERTa-Large model takes, on average in a GPU environment, 34ms per segment across all three datasets. The same BERTScore with TinyBERT takes 15ms per segment. Considering the practical examples 
made in the introduction (30k segments $\times$ 5 language pairs $\times$ 50 MT systems), the reference BERTScore would take, as we stated before, approx.\ 71 hours for a full pass, producing 8\cotwoeq of carbon footprint. Our suggested alternative, TinyBERT, would take more than twice as little, 31 hours, producing 3.62kg CO\textsubscript{2}-eq. In \se{a} CPU environment, the difference becomes even more striking. The reference metric with RoBERTa-Large would take more than 950 hours for a single full pass, while TinyBERT would complete in just under 45 hours. The carbon footprint is 5.4\cotwoeq for RoBERTa-Large and 0.26\cotwoeq for the TinyBERT-based metric (while maintaining 97\% of the reference model's quality). 

For those calculations, we assume that a GPU environment (Nvidia A100) has a power draw of 300W under full load, while a CPU environment has a power draw of 15W. However, the small size of TinyBERT would allow practitioners to use lower-tier GPUs, including some mobile GPUs in laptops, which are much more energy-efficient. For carbon footprint, we take the US 2021 power grid carbon intensity\footnote{\url{https://www.eia.gov/tools/faqs/faq.php?id=74&t=11}} as a reference point.

\paragraph{How stable are the results?}

We correlate the stability of (normalized) results using Pearson correlation over vectors corresponding to (a) metrics and (b) datasets. Concerning metrics, we observe high Pearson correlations from 0.72 (MoverScore-BaryScore, Quality) to 0.99 (MoverScore-BERTScore, inference time). This means that the LMs perform similarly (in terms of quality, inference time, or both) for each metric. Across datasets, the correlation is high for WMT15-WMT16 (0.91-0.99 for quality, runtime, and average) but considerably lower  between WMT21 and the other two (0.20-0.96). This is mainly because quality is not stable: for example, DeeBERT performs badly for WMT21, but quite well for WMT15 and WMT16. We note, however, that the score ranges for WMT21 are low anyway, and DeeBERT absolutely does not perform much lower than the other encoders here.

\paragraph{Which adapters are better?}
\se{For} both COMET and COMETINHO, 
the `parallel' adapter configuration consistently outperforms others, achieving high Kendall $\tau$ correlation values (0.289 for COMET, 0.254 for COMETINHO) and balancing forward and backward pass speeds. COMETINHO's `parallel' configuration also exhibits a 34\% reduction in memory usage and a remarkable backward pass speed.

Though it remains unclear why adapter-trained models surpass standard models in terms of quality, training curves are often close and sometimes overlapping. The \emph{parallel} adapter model achieves higher Kendall $\tau$ values despite reduced trainable parameters, which aligns with recent findings in~\citet{nouriborji-etal-2023-minialbert}.

\sen{Our} experiments utilize reference model hyperparameters, except for batch size and learning rate. Thus, further hyperparameter optimization might produce even higher correlations with human assessments.
\section{Conclusion}
We have investigated efficient evaluation metrics for natural language generation, particularly MT, in monolingual reference-based and multilingual reference-free versions via three approaches: (i) replacing transformers in metrics by efficient variants, (ii) replacing alignment models in metrics (precisely: Word Mover Distance) with efficient approximations, (iii) training COMET and COMETINHO \se{metrics} in a parameter-efficient way with adapters.
We have explored multiple types of efficient transformers, finding that TinyBERT shows the best quality-efficiency tradeoff for semantic similarity-based metrics: on average, it retains 97\% quality while being more than 5x faster at inference time and having considerably fewer parameters and lower memory consumption. In several cases, we have also identified faster models that yield higher quality at the same time. Finally, we found that efficient alignments on top of transformers do not result in efficiency gains but have adverse effects on quality in 2 out of the 3 datasets we examined. 

Our experiments \se{further} demonstrate that the `parallel' adapter configuration consistently outperforms others in efficiency and metric quality for both COMET and COMETINHO. The adapter-trained models achieve faster results, using less memory, and requiring a smaller portion of the reference model's training data, with COMET-sized adapter models being 
around 30\% faster to fully train (3h30m vs. 5h4m). These findings indicate that adapter-based training offers a promising approach for natural language generation tasks, providing optimal memory usage, computational efficiency, and alignment with human assessment.

In future work, we want to combine efficiency with other highly desirable properties of evaluation metrics such as \emph{robustness} \citep{Vu2022LayerOR,Chen2022MENLIRE,rony-etal-2022-rome} and \emph{explainability} \citep{kaster_global_2021,sai-etal-2021-perturbation,fomicheva_eval4nlp_2021,Leiter2022TowardsEE} to induce metrics that jointly satisfy these criteria.

\section*{Acknowledgements}
The Natural Language Learning Group is funded by the BMBF project <<Metrics4NLG>> and the DFG Heisenberg grant EG 375/5-1.

\section{Limitations}
In this section, we acknowledge several limitations of our research. First, with respect to WMD Approximations, we observe a surprisingly big drop in quality, on 2 out of 3 datasets, compared to the exact version. Thus, we cannot rule out that we made a mistake in the implementation. Nevertheless, our overarching conclusions remain valid since the performance improvements achieved by the WMD approximations are negligible in comparison to the time spent calculating the contextualized embeddings. Second, the experiments related to training efficiency with adapters reported in this paper focus primarily on the model's performance on the WMT2021 \textit{newstest} test set, which includes predominantly high-resource language pairs. Consequently, the results obtained in this study may not necessarily extend to other datasets or lower-resource language pairs. Lastly, the experiments with trained metrics were conducted using default adapter configurations and original hyperparameters from respective papers, including those of COMET and COMETINHO. A more comprehensive hyperparameter search could potentially improve metric quality and training efficiency even further. In particular, our models demonstrate superior performance compared to reference COMET and COMETINHO, albeit being trained on smaller datasets, suggesting that either the COMET's \& COMETINHO's hyperparameter configurations might not be the best one or that they were over-trained with more data than needed.
\bibliography{
custom,new}
\bibliographystyle{acl_natbib}

\appendix

\clearpage
\section{Related Work}
\label{appendix:related-work}
\textbf{Evaluation metrics}
Recent years have seen a surge of interest in transformer-based evaluation metrics as these promise higher quality (as measured in correlation with human assessments). Among the first \sen{LLM} based metrics are BERTScore \citep{zhang_bertscore_2020} and MoverScore \citep{zhao_moverscore_2019} which model evaluation as a semantic similarity task using contextualized BERT representations. Extensions include BARTScore \citep{yuan_bartscore_2021}, which reads off probability estimates as metric scores  directly from text generation systems, and MENLI \citep{Chen2022MENLIRE}, which builds on the paradigm of natural language inference and targets metric robustness. A distinction of evaluation metrics is whether they use human references (in \emph{reference-based} metrics) or do not use them (in \emph{reference-free} metrics), where the latter are less costly. In this work, we consider XMoverScore \citep{zhao-etal-2020-limitations} and SentSim \citep{song_sentsim_2021} as reference-free metrics and BARTScore, BERTScore, MoverScore and BaryScore \citep{colombo_automatic_2021} as reference-based metrics. In MT, the challenge for reference-free metrics is cross-lingual representation spaces, for which there are also different efficient transformer variants. Another distinction of metrics is whether they are trained or untrained. 
\textbf{Trainable/trained metrics} such as COMET \citep{rei_comet-22_2022}, which are particularly popular in the MT community, address the task of NLG evaluation by directly fine-tuning models on human-sourced annotation scores. COMET is trained on direct assessment scores given by human annotators to translations produced by various MT systems. As a result, COMET outperforms untrained metrics in terms of correlation with human assessment. However, the training process can be expensive due to increased model and data size.\footnote{A different approach to a text evaluation is presented in the recent papers of \citet{fu2023gptscore,liu2023gpteval}, which form a novel class of LLM based metrics. Those metrics rely on LLM predictions to assess different aspects of the text, with GPTScore \citep{fu2023gptscore} relying on token probabilities, while GPTEval \citep{liu2023gpteval} uses Chain-of-Thought to prompt the model to generate scores. }

\textbf{Efficiency} is a core issue of modern deep learning systems, which have become bigger and bigger in a quest for better performance, leading to environmental concerns and increasing inequality/exclusion. There are many approaches for obtaining more efficient models, especially in the context of computation-heavy transformers. 

\textbf{Knowledge Distillation}
\label{sec:knowledge-distillation} 
involves (i) a \textit{student} with pruned layers, embedding size or attention heads or even with layers replaced by alternative simpler network architectures and (ii) a \textit{teacher} usually of the same architecture, but larger. Then the student is trained with outputs from the teacher \citep{hinton_distilling_2015, ganesh_compressing_2021}. 
While data for traditional training often is scarce, in knowledge distillation, the teacher \textit{generates} new data for training the student \citep{hinton_distilling_2015}. Knowledge distillation thus usually leads to a better quality for the student model than traditional training because it benefits from the knowledge of the larger model. For example, DistilBERT \citep{sanh_distilbert_2020} consists of only 6 layers, which is half the amount of \bertbase{}, but with reported small reductions of model quality. TinyBERT \citep{jiao_tinybert_2020} introduces an improved method for knowledge distillation in which the student is trained with outputs (as DistilBERT does) and intermediate results.

\citet{kamal-eddine-etal-2022-frugalscore} apply knowledge distillation to induce an evaluation metric which they call FrugalScore.
They employ knowledge distillation on a set of pre-trained miniature BERT models, which were fine-tuned using a synthetic dataset created with full-size models employing BERTScore and MoverScore metrics.
In contrast to their approach, we much more comprehensively analyze efficiency in evaluation metrics.
Especially we consider more datasets and metrics, explore different evaluation environments (CPU-only and GPU), and improve training efficiency through adapters.

\textbf{Dynamic Inference Acceleration}:  
\citet{sun_patient_2019, xin_deebert_2020, zhu_leebert_2021} compare intermediate results after each encoder layer, after which they add an early exit ramp. If the change of the representation from one layer to the next is lower than a threshold, they assume later layers will improve even less and skip them by taking the early exit ramp. The final output for that sample will then be the representation from that layer.

\textbf{Adapters} is an approach that addresses the problem of training-time efficiency for fine-tuning transformer models. Popularized by \citet{pfeiffer2020AdapterHub,houlsby2019parameter} in the NLP community, adapters propose to freeze all parameters of the pre-trained transformer model and instead train adapters --- small intermediate layers added into the model graph. This reduces required memory usage because the gradients are computed only for a small number of parameters, though it is still required to keep the original model in memory, and can also lead to improvements in training speed. There are multiple ways to apply an adapter to the model. Simpler adapters~\citep{houlsby2019parameter,pfeiffer_mad-x_2020,he_towards_2022} offer a straightforward architecture with minimal modifications, focusing on training time and memory efficiency, while more complex ones~\citep{hu_lora_2021,mahabadi_compacter_2021,liu2022fewshot} provide the potential for better performance improvements at the cost of increased architecture complexity and resource consumption. For instance, the recently introduced $(IA)^3$~\citep{liu2022fewshot} adapter architecture is specifically designed to maximize fine-tuned accuracy after a few training steps. Another example is the method called UniPELT~\citep{mao-etal-2022-unipelt}, which combines prefix-tuning, low-rank adaptation (LoRA), and adapter layers.

\section{WMT Data}
\label{appendix:wmt-data}
\begin{table}[htb]
\centering
\begin{tabular}{cccc}
        \toprule
        \thd{pair} & \thd{WMT15} & \thd{WMT16} & \thd{WMT21} \\
        \midrule
        cs-en &  500 &  560 & \\
        de-en &  500 &  560 & \\
        fi-en &  500 &  560 & \\
        ro-en &      &  560 & \\
        ru-en &  500 &  560 & \\
        tr-en &      &  560 & \\
        zh-en &      &      & 3250 \\
        \midrule
        total & 2000 & 3360 & 3250 \\
        \bottomrule
    \end{tabular}
    \caption{Overview of the amount of segments per language pair in each dataset.}
    \label{appendix:tab:wmt-amount-segments}
\end{table}

\section{Experiment details}
\label{appendix:experiment-details}

\begin{figure*}
   \centering
    \begin{tikzpicture}
        \scriptsize
        \begin{axis}[
            axis lines=middle,
            xmin=0, xmax=420,
            ymin=0.345, ymax=0.46,
            legend pos = south west,
            legend cell align={left},
            grid = major,
            x label style={at={(axis description cs:0.5,-0.05)},anchor=north},
            y label style={at={(axis description cs:-0.05,0.5)},rotate=90,anchor=south},
            xlabel = runtime in ms/segment,
            ylabel = Pearson's r correlation with human evaluation,
            width=\linewidth, height=0.5\linewidth
        ]
            \addlegendimage{only marks,color=red}
            \addlegendentry{XMoverScore}
            \chartbase{ 405,0.4136}{red}{mBERT           }{180}{\srcshape}
            \chartscat{ 369,0.3485}{red}{\xlmrb{}        }{ 90}{\srcshape}
            \chartscat{ 186,0.2331}{red}{\xdistil        }{  0}{\srcshape}
            \chartscat{ 191,0.4375}{red}{\mminilmsix{}   }{  0}{\srcshape}
            \chartscat{ 234,0.4485}{red}{\mminilmtwelve{}}{  0}{\srcshape}

            \addlegendimage{only marks,color=cyan}
            \addlegendentry{SentSim}
            \chartbase{ 308,0.4235}{cyan}{\xlmrb{}        }{  0}{\srcshape}
            \chartscat{ 311,0.3948}{cyan}{mBERT           }{  0}{\srcshape}
            \chartscat{ 253,0.3899}{cyan}{DistilMBERT     }{315}{\srcshape}
            \chartscat{ 215,0.1238}{cyan}{\xdistil{}      }{  0}{\srcshape}
            \chartscat{ 210,0.4301}{cyan}{\mminilmsix{}   }{  0}{\srcshape}
            \chartscat{ 232,0.4554}{cyan}{\mminilmtwelve{}}{  0}{\srcshape}
        \end{axis}
    \end{tikzpicture}
    \caption{CPU runtime/correlation plot of XMoverScore and SentSim}
    \label{fig:plot-cpu-runtime-xs}
\end{figure*}
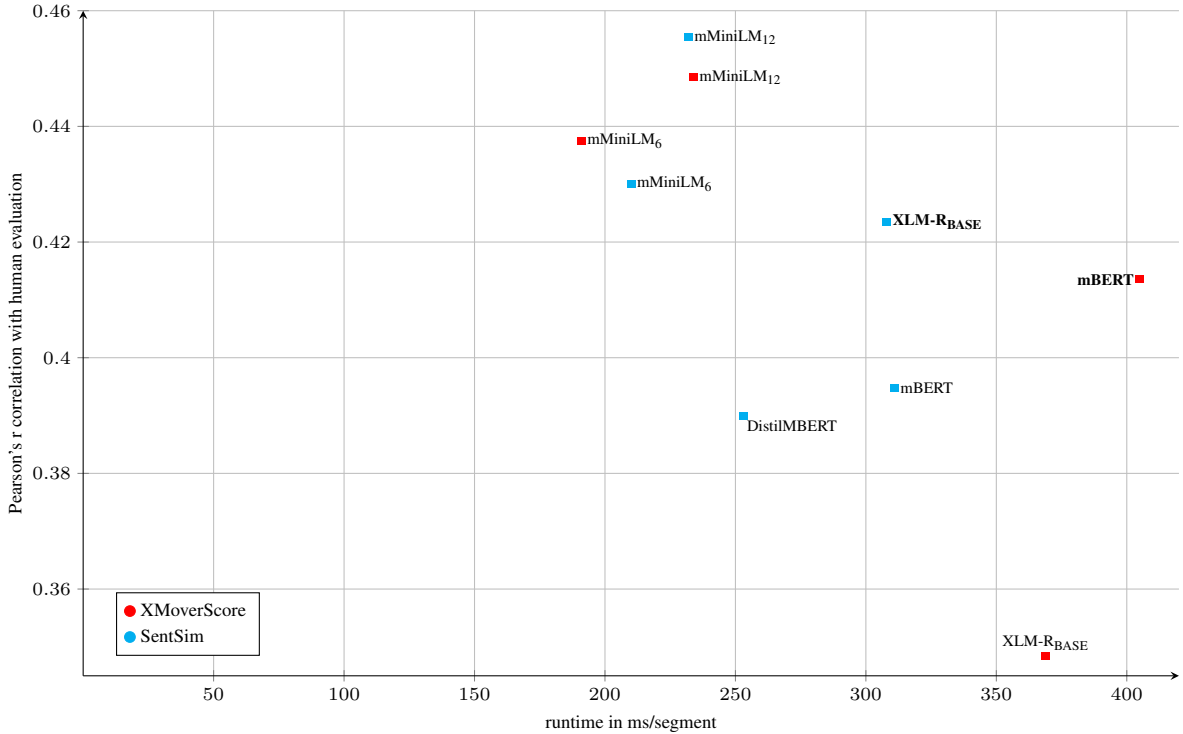

\begin{figure*}
        \centering
        \begin{subfigure}[c]{0.49\textwidth}
            \caption{CPU runtime}
            \label{fig:comparison-cpu}
            \begin{tikzpicture}
                \scriptsize
                \begin{axis}[
                    axis lines=middle,
                    xmin=0, xmax=430,
                    ymin=0.48, ymax=0.59,
                    legend pos = south east,
                    legend cell align={left},
                    grid = major,
                    x label style={at={(axis description cs:0.5,-0.05)},anchor=north},
                    y label style={at={(axis description cs:-0.1,0.5)},rotate=90,anchor=south},
                    xlabel = runtime in ms/segment,
                    ylabel = Pearson's r correlation with human evaluation,
                    width=\linewidth, height=\linewidth
                ]
                    \addlegendimage{only marks,color=blue}
                    \addlegendentry{BERTScore}
                    \chartbase{ 422,0.5856}{blue}{\robertalarge}{180}{\refshape}
                    \chartscat{ 137,0.5524}{blue}{\bertbase    }{315}{\refshape}
                    \chartscat{10.3,0.4823}{blue}{\berttiny    }{  0}{\refshape}
                    \chartscat{71.8,0.5428}{blue}{DistilBERT   }{270}{\refshape}
                    \chartscat{15.8,0.5300}{blue}{\tinybertfour}{315}{\refshape}

                    \addlegendimage{only marks,color=green}
                    \addlegendentry{MoverScore}
                    \chartbase{ 240,0.5412}{green}{\bertbase    }{  0}{\refshape}
                    \chartscat{20.5,0.4927}{green}{\berttiny    }{ 45}{\refshape}
                    \chartscat{ 121,0.5427}{green}{DistilBERT   }{  0}{\refshape}
                    \chartscat{34.0,0.5484}{green}{\tinybertfour}{ 45}{\refshape}

                    \addlegendimage{only marks,color=teal}
                    \addlegendentry{Ensembles}
                    \chartscat{ 131,0.5554}{teal}{A}{  0}{\refshape}
                    \chartscat{ 146,0.5657}{teal}{B}{315}{\refshape}
                    \chartscat{ 164,0.5666}{teal}{C}{  0}{\refshape}

                    \addlegendimage{only marks,color=gray}
                    \addlegendentry{FrugalScore}
                    \chartscat{21.9,0.4918}{gray}{fs-tiny-r   }{315}{\refshape}
                    \chartscat{ 236,0.5477}{gray}{fs-small-r  }{  0}{\refshape}
                    \chartscat{ 404,0.5807}{gray}{fs-medium-d }{225}{\refshape}
                \end{axis}
            \end{tikzpicture}
        \end{subfigure}
        \begin{subfigure}[c]{0.49\textwidth}
            \caption{GPU runtime}
            \label{fig:comparison-gpu}
            \begin{tikzpicture}
                \scriptsize
                \begin{axis}[
                    axis lines=middle,
                    xmin=0, xmax=46,
                    ymin=0.48, ymax=0.59,
                    legend pos = south east,
                    legend cell align={left},
                    grid = major,
                    x label style={at={(axis description cs:0.5,-0.05)},anchor=north},
                    y label style={at={(axis description cs:-0.1,0.5)},rotate=90,anchor=south},
                    xlabel = runtime in ms/segment,
                    ylabel = Pearson's r correlation with human evaluation,
                    width=\linewidth, height=\linewidth
                ]
                    \chartbase{34.1,0.5857}{blue}{\robertalarge}{  0}{\refshape}
                    \chartscat{22.8,0.5524}{blue}{\bertbase    }{  0}{\refshape}
                    \chartscat{ 7.8,0.4823}{blue}{\berttiny    }{  0}{\refshape}
                    \chartscat{16.9,0.5429}{blue}{DistilBERT   }{  0}{\refshape}
                    \chartscat{15.3,0.5300}{blue}{\tinybertfour}{  0}{\refshape}

                    \chartbase{45.3,0.5412}{green}{\bertbase    }{225}{\refshape}
                    \chartscat{28.2,0.4927}{green}{\berttiny    }{  0}{\refshape}
                    \chartscat{37.3,0.5428}{green}{DistilBERT   }{225}{\refshape}
                    \chartscat{33.0,0.5484}{green}{\tinybertfour}{ 45}{\refshape}


                    \chartscat{ 7.2,0.4918}{gray}{fs-tiny-r   }{  0}{\refshape}
                    \chartscat{ 8.8,0.5477}{gray}{fs-small-r  }{  0}{\refshape}
                    \chartscat{12.7,0.5807}{gray}{fs-medium-d }{  0}{\refshape}
                \end{axis}
            \end{tikzpicture}
        \end{subfigure}
        \caption{Selected metrics in runtime / correlation plots on a CPU and GPU.}
        \label{fig:comparison-cpu-gpu}
    \end{figure*}
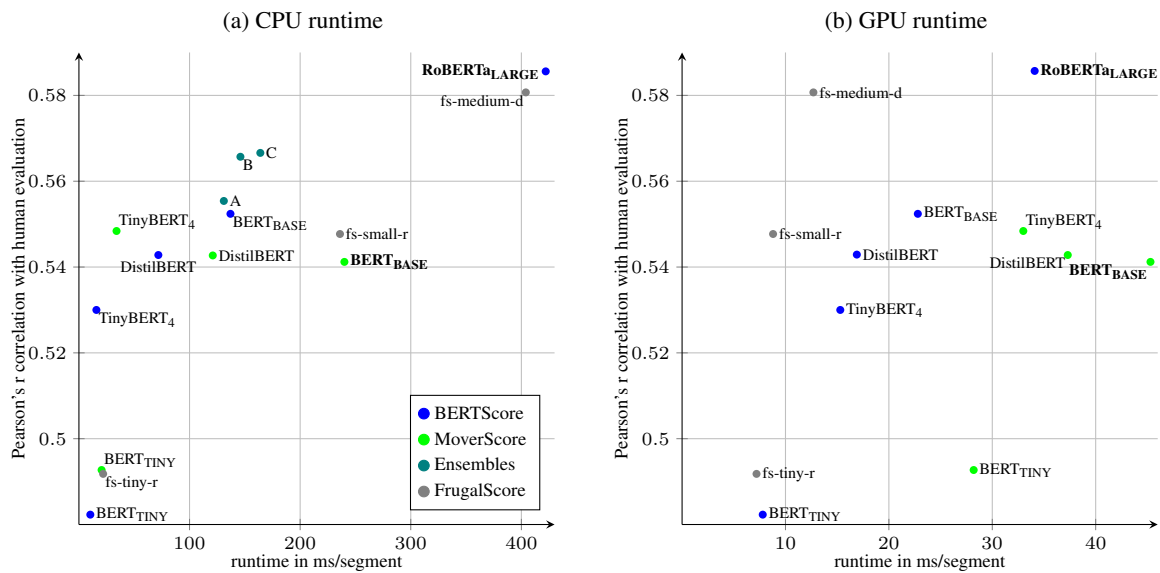

\subsection{
    BERTScore, MoverScore, BaryScore
}
\label{sec:appendix-eva-replace-models}
\paragraph{Setup}
Although BERTScore was built using various models, for comparing sentences in English (as both our references and hypotheses are), the authors suggest using \robertalarge{} \citep{zhang_bertscore_2020}. We only use it on BERTScore, since it was too slow with MoverScore and BaryScore. Its configuration is: $L=24, H=1024, A=16$ \citep{liu_roberta_2019}, where $L$ is the number of layers, $H$ the size of each layer and $A$ stands for the attention heads.
\custompar{\bertbase{}} is the original model of MoverScore and BaryScore \citep{zhao_moverscore_2019, colombo_automatic_2021} and also a possible optimization for BERTScore. It has $L=12$, $H=768$ and $A=12$~\citep{devlin_bert_2019}.
\custompar{\berttiny{}} is the smallest variant of BERT \citep{turc_well-read_2019} and was trained in the traditional way (directly on data, no Knowledge Distillation). It has $L=2$, $H=128$, $A=2$.
\custompar{DistilBERT} is a distillation of \bertbase{}. It has the same hidden dimensions of 768 and 12 attention heads, but only $L=6$. Also, they optimized the final output layers~\citep{sanh_distilbert_2020}.
\custompar{\tinybertfour{}} is another distillation of \bertbase{} with a more robust training method. It has $L=4$, $H=312$, and $A=12$~\citep{jiao_tinybert_2020}.
\custompar{\deebertmnli} is an early exiting version of \bertbase{}. It has the same structure, but after each encoder layer there is one added classification layer, that can be used as an off-ramp, to stop inference at intermediate states \citep{xin_deebert_2020}.

\paragraph{Results} We observe a speedup of the runtime that coarsely correlates with the size of the model. The figure shows that the fastest model on each of the metrics is \berttiny{}, which is up to 41x faster than the baseline (on BERTScore), but its quality also decreases by over 10 points correlation for BERTScore. A better speedup-quality ratio achieves \tinybertfour{}: The quality decreases by less than 6 points across all metrics (improves even for MoverScore). Furthermore, it is still up to 27x faster than the baseline (on BERTScore).  Memory measurements show similar behavior to runtimes, and they coarsely correlate with model size. \berttiny{} uses 61x and \tinybertfour{} 18x less memory than the baseline on BERTScore. Compared to \tinybertfour{}, DistilBERT shows  lower speedup and memory saving, but also a lower quality decrease (on all three metrics approximately half the decrease of \tinybertfour{}).
In our experiments, DeeBERT behaves very similar to \bertbase{}, both in quality and efficiency.

\subsection{BARTScore}
\label{sec:bartscore-appendix}
\paragraph{Setup} \custompar{\bartlargepara} is the original model used in BARTScore, fine-tuned on the Parabank2 dataset. It consists of 12 encoder and 12 decoder layers and has a hidden size of $H=1024$ \citep{lewis_bart_2019}. 
\custompar{\bartlargecnn} uses the same architecture, but after pre-training on CNNDM, it was not fine-tuned on Parabank2. We run experiments on \bartlargecnn{} and used these results as a baseline for fair comparisons to the other models. \custompar{\bartbase} was proposed by~\citet{lewis_bart_2019} and differs from \bartlarge{} in having 6 encoder and decoder layers instead of 12 and $H=768$.
\custompar{\dbartsix} is a distilled BART version with 6 encoder layers and 6 decoder layers. It was shrunk from \bartlarge{} and therefore has 
$H=1024$~\citep{shleifer_pre-trained_2020}.
\custompar{\dbarttwelve} is a distilled BART version with 12 encoder layers and 3 decoder layers. It was shrunk from \bartlarge{} and also has 
$H=1024$~\citep{shleifer_pre-trained_2020}.
\custompar{\dbarttts} has 12 encoder layers and 6 decoder layers and was trained on WikiSQL dataset. 
\custompar{\dbartmnli} has 12 encoder layers and 9 decoder layers, with $H=1024$.

\paragraph{Results} We observe differences among the baselines: \bartlargepara{} uses almost twice the memory of \bartlargecnn{}. The fastest model is again the smallest model -- \bartbase{} -- with 
speedups of 2.7x and 1.8x compared to \bartlargecnn{}. \bartbase{} is also the most memory-efficient model: with the usage of 663MB, it needs 2.6x less than \bartlargecnn{}. Despite being very efficient, \bartbase{}'s quality declines too much, with a correlation coefficient 12 points lower than the baseline (-25\%). Concerning quality, \dbartsix{} even gets a higher correlation than \bartlargecnn{} by 0.02 (+3.6\%). It brings a speedup of 1.8x (CPU) and 1.7x (GPU) and is 1.7x more memory efficient. \dbarttwelve{}, \dbarttts{} and \dbartmnli{} do not achieve competitive quality, also their acceleration over \bartlargecnn{} is only moderate. 

\subsection{Reference-free metrics}
\label{sec:xmoverscore-sentsim-word-model-appendix}
\paragraph{Setup}
To implement efficient reference-less metrics, we explore the following multilingual embedding models:
\custompar{mBERT} is the original model of XMoverScore used by \cite{zhao-etal-2020-limitations}. It has $L=12$, $H=768$, and $A=12$. It was trained on 104 languages \citep{devlin_bert_2019}. 
\custompar{\xlmrb} is the original model of SentSim used by \citet{song_sentsim_2021}. It has $L=12$, $H=768$ and $A=12$ and was trained on data in 100 languages \citep{conneau_unsupervised_2020}.
\custompar{DistilMBERT} is a distillation of mBERT \citep{sanh_distilbert_2020}. It has the same dimensions and attention heads, but only $L=6$ layers and the final output layers were stripped. These missing output layers are what makes this model incompatible to XMoverScore. Thus, we only use this model for SentSim. 
\custompar{\xdistil} is another distillation of mBERT \citep{mukherjee_xtremedistil_2020}. The model, called \textit{TinyMBERT} in the first version of the paper, has $L=6$, $H=256$ and $A=12$. 
\custompar{\mminilmsix} is a distillation of \xlmrl{} with $L=6$, $H=384$, and $A=12$.
\custompar{\mminilmtwelve} is a distillation of \xlmrl{} with $L=12$ \citep{wang_minilmv2_2021}.
{\bf For XMoverScore with language models} the two compared models are: 
\custompar{GPT-2} \citep{radford_language_2019} is the original model used by XMoverScore \citep{zhao-etal-2020-limitations}.
\custompar{DistilGPT-2} is a distillation of GPT-2 \citep{von_platen_distilgpt2_2021}.
{\bf For SenSim with sentence embeddings} we replace the original sentence embedding model with a lighter one. We try 3 other models from  the SBERT framework. For reasons of clarity, we abbreviate the names to a letter from the SBERT site: 
\custompar{A: xlm-r-bert-base-nli-stsb-mean-tokens}, the original model used by \cite{song_sentsim_2021}.
\custompar{B: distiluse-base-multilingual-cased-v2} - a DistilBERT-based model which was fine-tuned on synthetic data created with Universal Sentence Encoder~\citep{cer2018universal}, \custompar{C: paraphrase-multilingual-MiniLM-L12-v2} - based on MiniLM model with 12 layers, and 
\custompar{D: paraphrase-multilingual-mpnet-base-v2} - which is based on MPNet model~\citep{song2020mpnet}.

\paragraph{Results} We present the results in Figure~\ref{fig:plot-cpu-runtime-xs}. Both mMiniLMs outperform their baselines in quality and efficiency. The 6-layer version is up to 2.1x faster (on CPU) than the baseline and has a 0.024 higher correlation for XMoverScore (+5.8\%). Even higher is the quality improvement with \mminilmtwelve{}: it achieves a 0.032 higher correlation than the baseline on XMoverScore (+8.4\%). Both models also show quality improvements on SentSim. The space they occupy on a disk is $1/5$ of \xlmrb{} and $1/3$ of mBERT. Although the needed memory is a lot higher than the disk space, with up to 1,594MB (\mminilmtwelve{} on SentSim), the models still need 1.2x (SentSim) and 1.4x (XMoverScore) less inference time than the baselines. DistilMBERT on SentSim also shows speedups of 1.2x on both CPU and GPU but has a 0.034 lower Pearson's r than \xlmrb{}. \xdistil{}, despite bringing some memory efficiency and a big saving on disk space, has a quality that is too bad on both metrics. The remappings on XMoverScore do not bring any difference for mBERT and \xlmrb{}. For the mMiniLMs,  we observe a quality increase of 1.6 points (6 layers) and 1.3 points (12 layers) correlation compared to using no remapping. Only for \xdistil{}  do we see a real difference: using the remappings improves correlation by approximately 0.122 (+110\%) compared to using the model directly. 
For XMoverScore with language models we can not see any change in runtime on a CPU, but observe a speedup of 1.3x on a GPU. We also see a lower memory usage of 317MB (20\%) and 
a lower disk-space of a third. 
The 
quality drops by 
3.5 points.
For SentSim with sentence embedding models we observe a slight speedup of Model B of 1.2x (CPU) and 1.3x (GPU), but also a decrease of quality of 
5.5 points. 
Model C shows only small decrease of quality of 
1.7 points, 
but is very efficient: it runs in 1.5x faster speed on CPU and 1.1x faster on GPU, saves memory (1.4x less) and also only occupies 43\% of the disk-space. 
Model D 
achieves a slightly higher correlation 
(+1.2 points)
and has the same size as the baseline (comparable speed and memory).

\subsection{WMD}
\label{sec:wmd-appendix}
\paragraph{Results} We see a drop in quality from a correlation of 0.54 
on average over all three datasets to 0.43 and 0.39, but surprisingly, we see a substantial increase of quality when using RWMD on WMT21. On the other hand, the quality of XMoverScore, as indicated by the correlation with human scores, declines when using these more efficient variants. 
WCD achieves correlations approximately 10 percent lower than WMD. As on MoverScore, RWMD's correlations drop approximately 30 percent for WMT15 and WMT16. For WMT21, we again observe a very high correlation with RWMD.

\paragraph{Runtime} 
For neither MoverScore nor XMoverScore do we observe a substantial speedup while using WCD or RWMD instead of WMD. Thus, we investigate the time consumption of each calculation step in more detail.

\begin{table}[htb]
{\small
    \begin{tabularx}{\linewidth}{@{} l *3{|>{\raggedleft\arraybackslash}X} @{}}
        \toprule
        \thd{Step} & \thd{WMD} & \thd{WCD} & \thd{RWMD} \\
        \midrule
        get BERT embeddings       & 285.499 & 287.915 & 291.122 \\
        calculate distance matrix &   0.829 &   0.005 &   0.782 \\
        calculate distance        &   5.602 &   0.616 &   0.449 \\
        \bottomrule
    \end{tabularx}
    }
    \caption[Runtime of each step of MoverScore for various distance functions]{Runtime (in ms) of each step of MoverScore for various distance functions using its original model \bertbase{}.}
    \label{tab:moverscore-steps}
\end{table}

In Table \ref{tab:moverscore-steps}, we can see that the calculation of WCD and RWMD is substantially
faster than WMD --- 9x and 12x on MoverScore and 20x and 27x on XMoverScore. But it also shows that the calculation of the embeddings (and of the perplexity) takes a much longer time than the calculation of the distance. No matter how fast the distance can be calculated, the speedup will be eaten up by the variance of the calculation time for the embeddings (and perplexity).

\section{Impact of different batch sizes}
\label{appendix:batch-size}
In order to examine the effects of different batch sizes, we conducted reduced experiments with a smaller number of models, testing only on BERTScore and WMT15 dataset. See Table~\ref{tab:bs-ablation} for a list of hyperparameters.
\begin{table}[h!]
    \centering
    \begin{tabular}{c|c}
        Hyperparameter & Setting \\
        \midrule
        Model & \bertbase,
         \berttiny, \\ & \tinybertfour, DistilBERT \\
        Environment & CPU and GPU cluster nodes \\
        Batch size & 1, 4, 16, 64
    \end{tabular}
    \caption{Hypeparameter settings for batch-size ablation study.}
    \label{tab:bs-ablation}
\end{table}
As before, we report averaged results of 3 runs to account for fluctuations.

\begin{table}[h!]
    \centering
    \begin{tabular}{c|l|l|l|l|l}
        Model & 1 & 4 & 16 & 64 & $r$ \\
        \midrule
        \bertbase &     .084 & .021 & .010 & .009 & .729\\
        DistilBERT &    .032 & .012 & .006 & .005 & .709\\
        \tinybertfour & .015 & .015 & .003 & .002 & .707\\
        \berttiny &     .005 & .006 & .005 & .002 & .635\\
    \end{tabular}
    \caption{Runtime duration (seconds per segment) for BERTScore with given models and variable batch size in CPU environment. $r$ stands for Pearson's correlation with human judgement and provided for brevity.}
    \label{tab:bs-variation-cpu}
\end{table}

\begin{table}[h!]
    \centering
    \begin{tabular}{c|l|l|l|l|l}
        Model & 1 & 4 & 16 & 64 & $r$ \\
        \midrule
        \bertbase &     .021 & .006 & .003 & .002 & .729\\
        DistilBERT &    .013 & .004 & .002 & .002 & .709\\
        \tinybertfour & .011 & .004 & .002 & .001 & .707\\
        \berttiny &     .007 & .003 & .002 & .002 & .635\\
    \end{tabular}
    \caption{Runtime duration for BERTScore with given models and variable batch size in GPU environment. $r$ stands for Pearson's correlation with human judgement.}
    \label{tab:bs-variation-gpu}
\end{table}

According to the results in Tables~\ref{tab:bs-variation-cpu} and~\ref{tab:bs-variation-gpu} there is a notable change in the model's relative efficiency compared to each other with change of the batch size. nevertheless, the best tradeoff between metric's quality and efficiency is still provided by \tinybertfour~model. It is also worth noting that in the case of the GPU environment, we observe faster saturation of efficiency gains between models of different sizes. On higher batch sizes, they perform around the same. However, switching to larger batch sizes leads to progressively higher memory consumption.

\end{document}